\documentclass[10pt,notitlepage]{article}
\usepackage{algorithm}
\usepackage{algpseudocode}
\usepackage{varwidth}
\usepackage{amsmath,epsfig}
\usepackage{subfig}
\usepackage{enumitem}
\usepackage{balance}
\usepackage{fancyhdr}
\usepackage{doi}
\usepackage[top=1in, bottom=1.2in, left=1.in, right=1.in]{geometry}
\usepackage{multicol}
\usepackage{float}
\usepackage{wrapfig}
\graphicspath{{./images/}}
\DeclareGraphicsExtensions{.pdf,.jpeg, .png, jpg}
\usepackage[square,numbers]{natbib}
\newcommand*{\thead}[1]{\multicolumn{1}{c}{\bfseries #1}} 
\newcommand{\hide}[1]{}

\usepackage{titlesec}
\titleformat{\section}
{\normalfont\Large\bfseries}{\thesection}{0.3em}{}
\titleformat{\subsection}
{\normalfont\bfseries}{\thesubsection}{0.3em}{}

\date{}

\renewenvironment{abstract}{%
	\small

	\begin{center}%
		{\bfseries \abstractname\vspace{-0.1em}\vspace{0pt}}%
	\end{center}%
	\it
	\quotation
}
{\endquotation}
\begin{document}
	
	\title{\textbf{Partially fake it till you make it:\\ mixing real and fake thermal images for improved object detection}}
	
	\author{Francesco Bongini, Lorenzo Berlincioni, Marco Bertini, Alberto Del Bimbo\\
		MICC - Universit\`a degli Studi di Firenze
		Firenze
		Italy
		\\
	name.surname@unifi.it}
	
	\maketitle
	
	\begin{figure}[h]
		\centering
		\includegraphics[height=0.18\textheight]{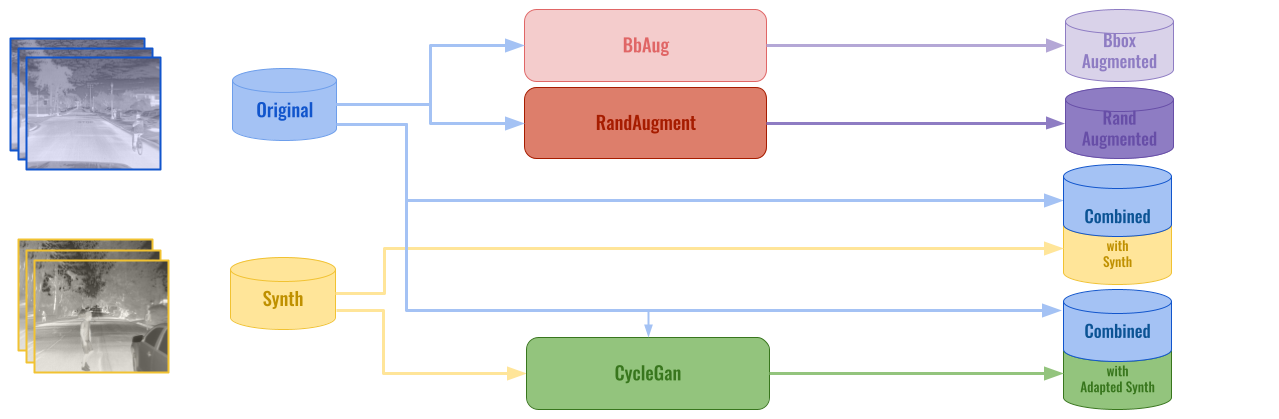}
		\caption{System overview: showing fake+real dataset augmentation, GAN augmentation, BBox augmentation.}
		\label{fig:teaser}
	\end{figure}

	\begin{multicols}{2}

	\begin{abstract}
		In this paper we propose a novel data augmentation approach for visual content domains that have scarce training datasets, compositing synthetic 3D objects within real scenes. We show the performance of the proposed system in the context of object detection in thermal videos, a domain where \textit{i)} training datasets are very limited compared to visible spectrum datasets and \textit{ii)} creating full realistic synthetic scenes is extremely cumbersome and expensive due to the difficulty in modeling the thermal properties of the materials of the scene. We compare different augmentation strategies, including state of the art approaches obtained through RL techniques, the injection of simulated data and the employment of a generative model, and study how to best combine our proposed augmentation with these other techniques.
		Experimental results demonstrate the effectiveness of our approach, and our single-modality detector achieves state-of-the-art results on the FLIR ADAS dataset.
	\end{abstract}


	\section{Introduction}
\begin{figure}[H]
		\centering
		\includegraphics[width=0.44\linewidth]{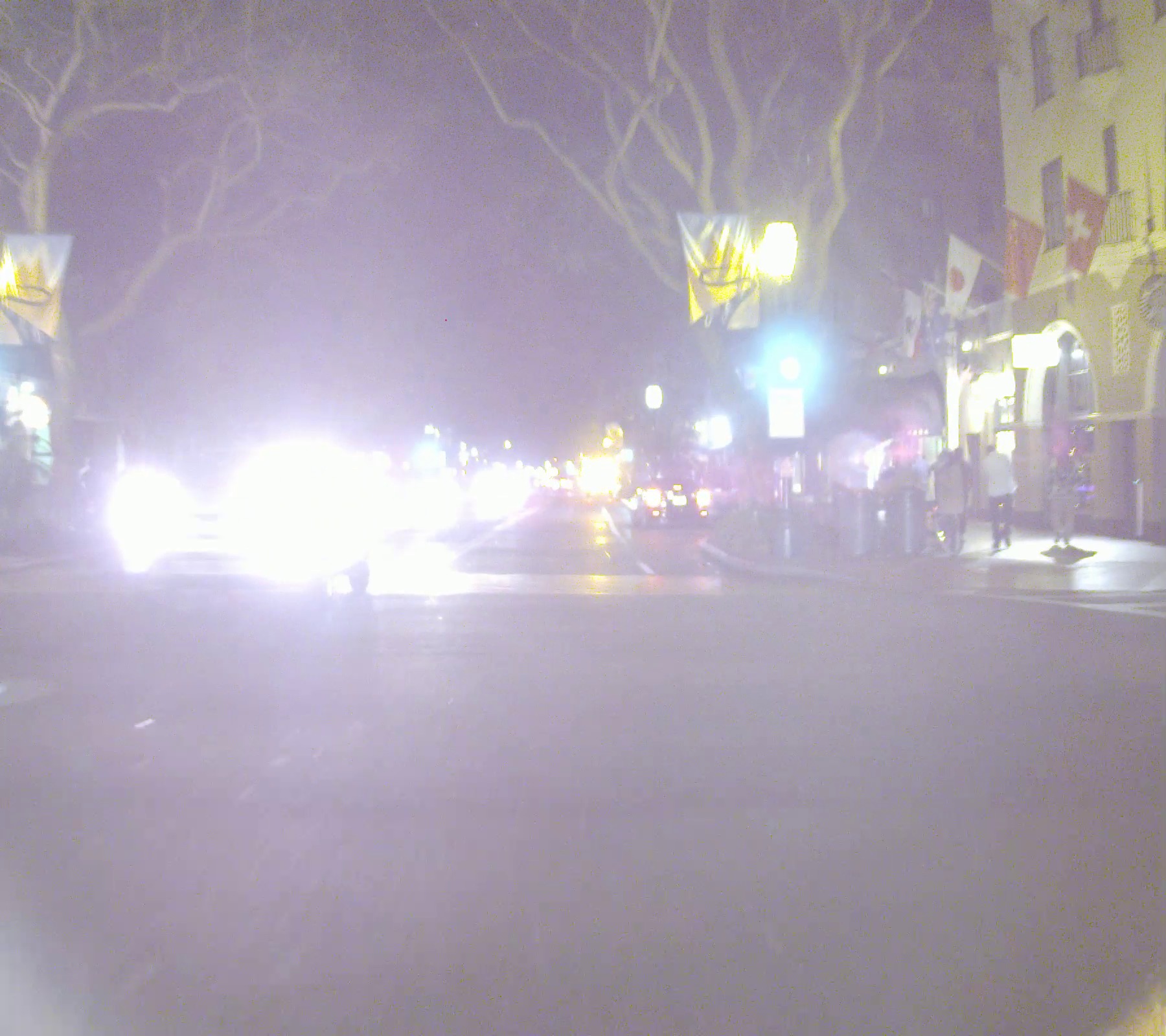}
		\includegraphics[width=0.49\linewidth]{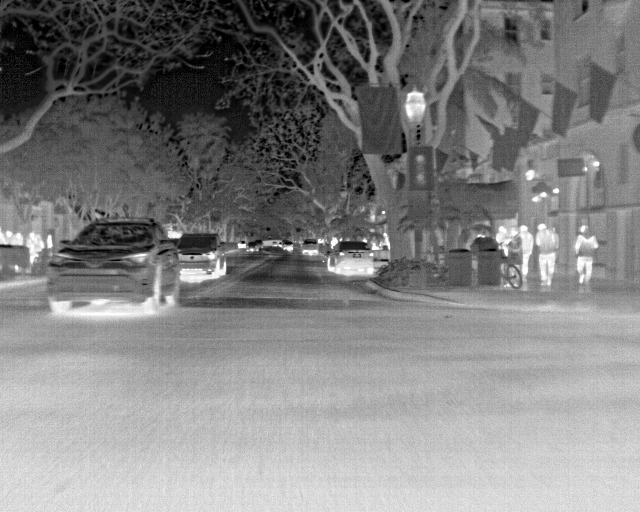}	
		\includegraphics[width=0.94\linewidth]{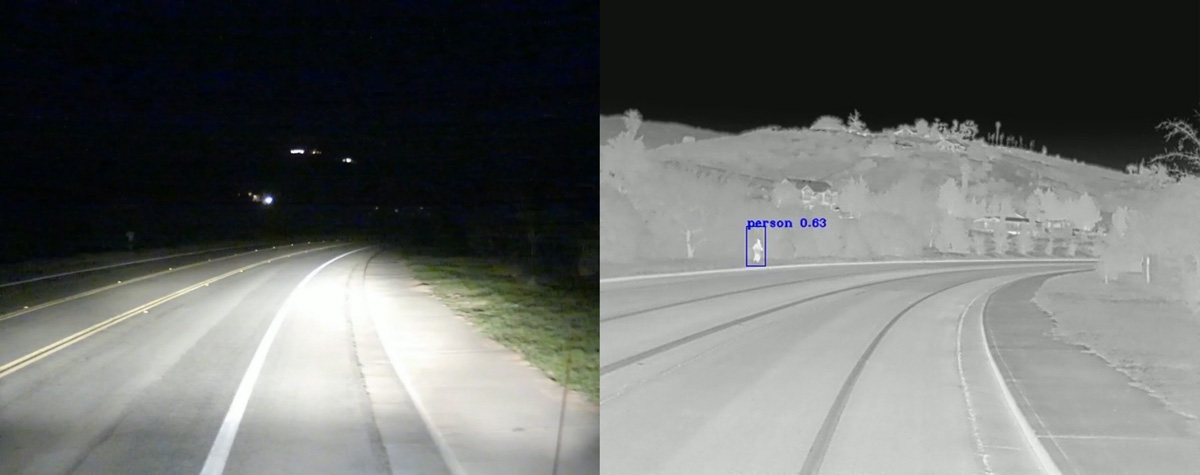}
		\caption{Top row. \textit{Left)} The lights of the car headlights dazzle the RGB camera; \textit{right)} cars and persons can be recognized in the same view obtained through a thermal camera.\\
			Bottom row. \textit{Left)} Lack of lighting may cause dangerous situations, no pedestrian is observable in the scene. \textit{Right)} the thermal camera does not depend from the intensity of the light and allows to detect the person.} \label{fig:rgb_thermal}
	\end{figure}

	Object detection is a core problem for the perception capabilities of an autonomous vehicle, the identification of its surroundings and of nearby objects is essential to ensure a safe deployment of autonomous cars on the road. 
	In autonomous driving the object detection task is required to be particularly robust across a range of illumination and environmental conditions, including daytime, nighttime, rain, fog, etc. In such conditions, detectors based solely on visible spectrum imagery can easily fail~\cite{li2018multispectral,kieu2019domain}, as shown in Fig.~\ref{fig:rgb_thermal}.
	
	The use of thermal detectors has recently increased as a mean to mitigate the sensitivity of visible spectrum imagery to scene-incidental imaging conditions~\cite{kieu2019domain, herrmann2018cnn, baek2017efficient}. A growing number of works have also investigated multispectral detectors combining visible and thermal images for robust pedestrian detection~\cite{tian2015pedestrian,angelova2015real,wagner2016multispectral,konig2017fully,brazil2017illuminating,guan2019fusion,li2018multispectral,li2019illumination}.
	
	The cost of multiple aligned sensors however is a limiting factor due to cost, camera size or availability of lighting to operate the visible spectrum camera, and multispectral models can have limited applicability in real-world applications. Moreover, using visible spectrum sensors does not offer the same privacy-preserving affordances as systems employing only thermal sensors~\cite{kieu2019domain}.
	
	When compared to multispectral detectors, thermal-only ones typically reach lower performances, and besides the lack of the information coming from the visible spectrum there are other challenges for robust object detection using only thermal data. A key performance-limiting factor is the relative lack of annotated thermal imagery available for training state-of-the-art models. Thermal pedestrian datasets are few, and -- compared to visible-spectrum datasets -- have orders of magnitude fewer annotated instances.
	As an example the Caltech Pedestrian Dataset \cite{dollar2011pedestrian} has 350,000 annotations in the visible domain, while in thermal domain KAIST Multispectral Pedestrian dataset \cite{hwang2015multispectral} has $\sim 51,000$ annotations and FLIR ADAS Dataset \cite{flir-adas} has $\sim 28,000$. 
	Bringing thermal-only detection to the robustness and accuracy demanded by real-world applications is thus extremely difficult due to the scarcity of annotated data.
	
	Creating realistic training data from 3D models has become a common practice in several domains, e.g.~for Advanced Driver Assistance Systems (ADAS). An example is the CARLA simulator \cite{Dosovitskiy17} that can create RGB, LiDAR, depth, etc. views of a simulated scene. However, thermal imagery is not yet automatically generated, since the 3D engine should account for a very large number of simulation parameters associated to different materials and heat sources.
	
	In this work we propose to simplify the creation of training data adding only a subset of objects of interest to a real scene, thus reducing the cost of modeling a whole environment. In order to improve the quality and visual likelihood of the objects we use a GAN, to further adapt the appearance of the 3D models and to better simulate the output of the shaders used in the 3D animation engine.\\ 
	
	The contributions of this paper are:
	\begin{itemize}
		\item \textbf{Compositing of 3d fake thermal objects in a real thermal scene}: to the best of our knowledge we are the first to propose this approach, addressing the thermal spectrum domain, in which creating full virtual scenes is not practical. We further propose to use a GAN-based approach to adapt the appearance of the composited objects in the scene.
		\item \textbf{Extensive comparison of data augmentation techniques}: we provide an extensive comparison of recent augmentation techniques that have been originally proposed and evaluated on RGB datasets, like RandAugment \cite{randaug} and BBAug \cite{zoph2020learning}, on thermal data; we evaluate their combinations, either between them and with our 3D compositing approach.
		\item \textbf{State-of-the-art results}: we compare the best thermal augmentations, either alone or in combination, with a large number of multispectral and thermal object detectors. Our proposed augmentation technique outperform all the thermal-only approaches and all the multispectral methods but one.
	\end{itemize}

	\section{Related work}
	
	\subsection{Object detection in thermal imagery}
	
	Thanks to the reduction of costs and availability of multispectral cameras over the
	past few years, there are numerous recent works exploiting thermal images in
	combination with visible images for robust pedestrian detections~\cite{wagner2016multispectral,liu2016multispectral,konig2017fully,xu2017learning,li2018multispectral,li2019illumination,fritz2019generalization,zhang2019cross,cao2019box,vandersteegen2018real,lee2018pedestrian,zheng2019gfd}.
	One of the key issues  with thermal imagery is the image resolution and
	quality is typically far lower compared to RGB images, and the application of RGB detectors on this low-quality data yields lower performances. 
	More recently works as 
	MMTOD-UNIT \cite{devaguptapu2019borrow} use a multi-spectral detector that aims to borrow the knowledge from the data-rich domains such as visual (RGB) without the explicit need for a paired multimodal dataset. In fact, the model introduces pseudo-visible images generated from the thermal spectrum ones using the CycleGAN \cite{CycleGAN2017}, promoting more information to the training set. The multi-modal Faster-RCNN detector is then trained.
	Another recent multi spectral detector is CFRM\_3 \cite{zhang2020multispectral} that instead  cyclically fuse and refine more spectra. The idea is to use a novel cycle fuse-and-refine module to predict the segmentation mask features of both visible and thermal spectrum. This allows to complementary get important features from different spectra. Note that this method  uses only aligned pictures so this is not directly comparable with the other models, although multispectral cameras typically do not provide aligned frames due to differences in the focal length of the two lenses. In particular, 4,129 well-aligned image pairs have been used for training and 1,013 image pairs for test. 
	
	In contrast, many recent works have investigated pedestrian detection in the
	thermal (IR) domain only. For example, authors in~\cite{john2015pedestrian}
	used Adaptive fuzzy C-means for IR image segmentation and a CNN for pedestrian
	detection. In~\cite{baek2017efficient} the authors proposed a combination of
	Thermal Position Intensity Histogram of Oriented Gradients (TPIHOG) and the
	additive kernel SVM (AKSVM) for nighttime-only detection in thermal imagery.
	Thermal images augmented with saliency maps, used as attention mechanism, have
	been used in~\cite{ghose2019pedestrian}.
	
	The idea of performing several video preprocessing steps to make thermal images
	look more similar to grayscale images converted from RGB was investigated
	in~\cite{herrmann2018cnn}, who then applied a pretrained and fine-tuned SSD
	detector. Recently, authors in~\cite{9064036} designed dual-pass fusion block
	(DFB) and channel-wise enhance module (CEM) to improve the one-stage detector
	RefineDet, and proposed their ThermalDet detector for pedestrian detection in
	thermal imagery. Another recent single-modality work was the Bottom-up Domain
	Adaptation approach proposed in~\cite{kieu2019domain, kieu2020layerwise} for pedestrian detection
	in thermal imagery. Task-conditioned training has been recently proposed in \cite{eccv-2020}, adding the auxiliary task of classifying night and day thermal images and obtaining state-of-the-art results. Also in this work, we focus on the thermal-only detection problem.
	
	
	\subsection{Data augmentation from 3D models}
	Creation of virtual images, typically using 3D graphic engines, to improve object detection and pose estimation in visual spectrum domain has received a lot of attention from the research community. In the autonomous driving field in particular countless synthethic datasets and virtual environments \cite{richter2016playing,7780721,DBLP:journals/corr/GaidonWCV16, martinez2017beyond} are used, especially for reinforcement learning models.
	The advantages of using a virtual environment are multiple such as the ease of collection, the control over the generated data and the possibility to extract information from multiple, perfectly aligned, virtual sensors i.e.~in a 3d videogame the RGB image, its segmentation map and depth map \textit{come for free} from the render program. 
	Specific tools have been developed to ease the creation of these scenarios, like CARLA \cite{Dosovitskiy17}, that is specialized for the creation of scenarios for ADAS applications, and VIVID \cite{vivid-2018}, that can be used for indoor navigation, action recognition and event detection, and provides an advanced human skeleton system to simulate complex human actions.
	Most of the effort for realism in these tools, as well as in the industry of 3d videogame engines, is focused on the visible spectrum, and the simulated thermal views of a scene usually lack the nuances of its visible counterpart.
	
	Because of this, works that have explored the use of tools to create augmented datasets with 3D data have addressed only the visible spectrum context.
	Some works have used 3D data to estimate hand \cite{oberweger2016efficiently} or body poses \cite{chen2016synthesizing, Varol_2017_CVPR}.
	3D avatars have been recently used to enhance human action recognition in \cite{ludl2020enhancing, ballout2020benefits}.
	Object detectors have been trained using 3D syntethic data in \cite{bochinski2016training} and \cite{Di-Benedetto:2020wb}. The former considered classes like person, vehicles and animals in outdoor city scenes, evaluating the trained detector on surveillance videos. The latter investigates the effectiveness of rendering engines in generating realistic scenes for scenarios where no or insufficient annotated data is available, considering the detection of protective equipment in construction sites. 
	
	To the best of our knowledge we are the first to address compositing of 3D objects within real scenes in the thermal domain.

	\subsection {Data augmentation from synthetic images}
	The exploiting of synthetic data from a simulator to obtain a larger, more diverse, dataset has been explored in many works. This presents a domain adaptation challenge in which the source domain that needs to be adapted is partially controllable.
	One approach to this problem focuses on the model i.e.~by adding to it a domain classifier and using gradient reversal to close the gap between the two distributions~\cite{ganin2015unsupervised} or by freezing the feature extraction section of the model when training it on synthetic data \cite{kieu2020layerwise}.
	Another technique works on closing the domain shift by having a shared representation space \cite{hong2018conditional,sankaranarayanan2018learning,pmlr-v70-long17a}.
	Another possible approach instead separately preprocesses the input data using a learned generative model, trained to perform a translation between the two domains.
	\cite{shrivastava2017learning,BARTH2020105378,augGAN,Hoffman2018CyCADACA}.
	Also in~\cite{devaguptapu2019borrow} the authors used a CycleGAN~\cite{CycleGAN2017} for image-to-image translation of
	thermal to pseudo-RGB data. The use of these frameworks to perform data
	augmentation in order to improve the performance of a separate classifier has
	been studied in multiple previous works such as
	\cite{antoniuo-2018} in which they focus on improving one-shot
	learning, in \cite{Bowles2018GANAA} where segmentation of medical images is
	enhanced by GAN augmented data. In \cite{pan2017virtual} synthetic data coming from a simulator is adapted and used to train an RL agent for autonomous driving.
	Similar to our approach are \cite{Hoffman2018CyCADACA} in which a much more elaborate cycle consistent framework is developed to perform domain adaptation and \cite{mo2019instagan} that improves the CycleGAN approach by adding per 
	instance masks. 
	In \cite{kieu2021robust}, a GAN is used to produce fake thermal images to increase the training data; the authors study the best approach on how to mix these additional data to real thermal images, showing how adding 10\%-20\% percent of fake images improves the performance of the object detection.

	\subsection{Data augmentation from RGB images}
	AutoAugment \cite{autoaug} is a augmentation framework for vision models that casts the search of parameters for data augmentation as an optimization problem and solve it using reinforcement learning. RandAugment \cite{randaug} improves on AutoAugment \cite{autoaug} by both considerably reducing the parameters search space from $10^{32}$ to $10^{2}$ and matching or exceeding performances of \cite{autoaug}. Both this models are tailored to image classification tasks, more recently a number of works focused on data augmentation  specifically developed for detection problems, such as \cite{ning2021data} with a neural rendering approach and \cite{zoph2020learning} using a similar technique to \cite{randaug} the authors train a RNN model in a RL setting to learn optimal augmentation policies including in this transformations translating, zooming or distorting both the image and the bounding boxes.

	\section{The proposed method}
	\subsection{Mixing real and fake thermal imagery}
	
\begin{figure}[H]
		\includegraphics[width=0.48\columnwidth]{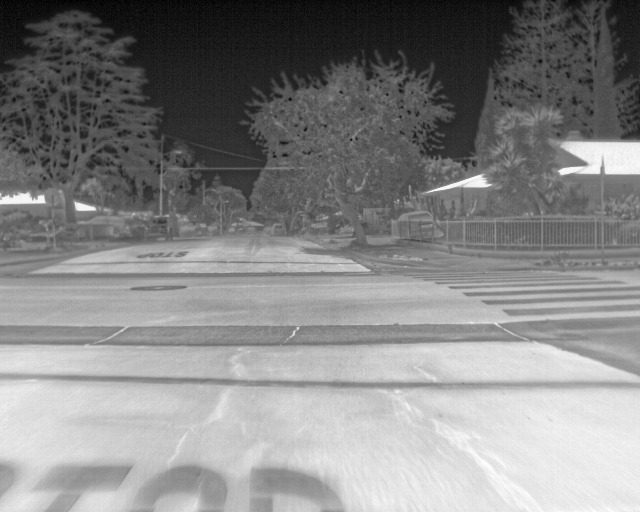}	
		\includegraphics[width=0.48\columnwidth]{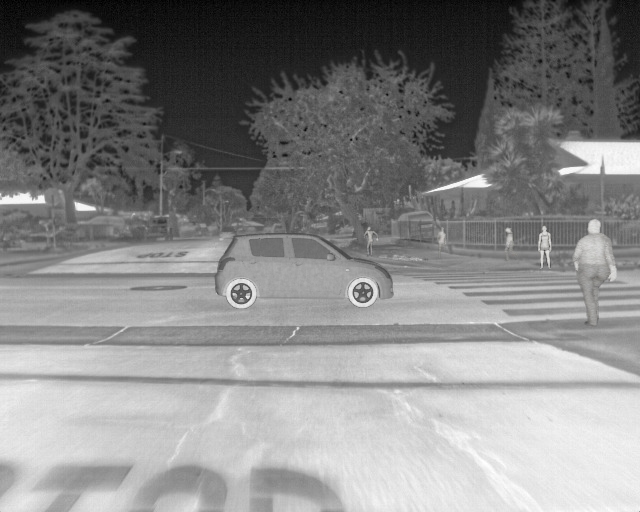}
		\caption{\textit{Left)} Original thermal image from FLIR dataset \cite{flir-adas}, \textit{Right)} Image with composited fake thermal objects.} \label{fig:scenes_1}
	\end{figure}

	The Unity game engine \cite{unity3d} is a powerful framework used to develop modern 3D games. With the help of a 3D artist we exploit this tool to get synthetic data by rendering sequences of realistic 3D characters and cars superimposed over a real thermal spectrum image as background.
	Fig.~\ref{fig:scenes_1} shows an example of the process: a real scene from FLIR dataset is used to composite and animate 3D objects, creating a virtual video sequence. The game engine allowed the use of a raw \textit{thermal} shader that simulates the thermal signature of different objects. The 3D artist needs to concentrate only on creating the raw thermal appearance of these objects, without need to design the whole scene. This would have required to compute the thermal properties of all the objects in the background.
	
	We automatically annotated the objects and collected over 10K samples used to augment the training set. In order to better analyze the synthetic data effect on the training we produced different sets of simulated scenes changing parameters such as the source of the background and the number of instances of the different detected categories (Person, Car, Bicycle). 
			\begin{figure}[H]
		\includegraphics[width=0.49\linewidth]{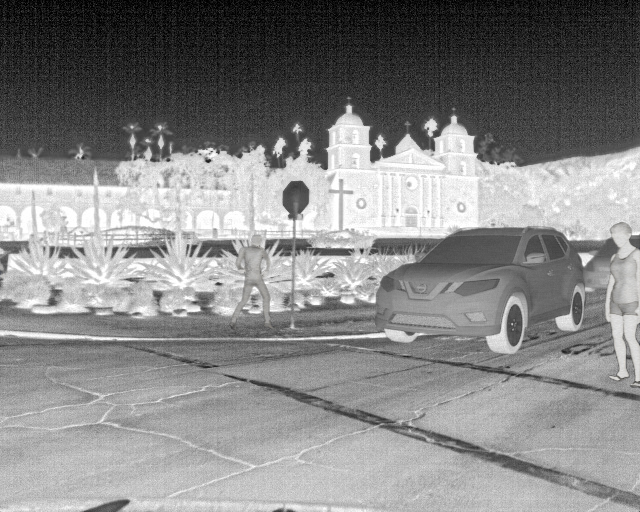}	
		\includegraphics[width=0.49\linewidth]{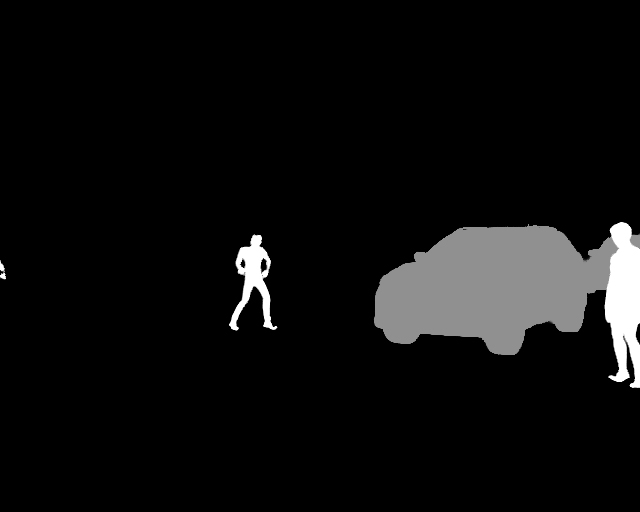}
		\caption{GAN Input data sample: \textit{left)} composited 3D fake thermal objects in a scene; \textit{right)} segmentation masks of the 3D objects $(img,seg)$ } \label{scenes_2}
	\end{figure}
	\subsection{Improving fake thermal imagery appearance through GAN}
	We use a Generative Adversarial Network to improve the realism of the data coming from the simulator.
	In our approach we rely on an LSGAN objective trained in a cycle-consistent regime as in \cite{CycleGAN2017}.  In order to achieve this we trained the generative model in a cycle-consistent setting between the \textit{source} domain, the synthetic data, and the \textit{target} domain, the data from FLIR dataset. The model architecture is taken from \cite{kieu2021robust} as it proved to be successful in the thermal images augmentation setting and it is built using the Residual in Residual
	Dense Block (RRDB) as the fundamental unit.
	As in \cite{Lim2017EnhancedDR}, we remove the batch normalization layer from the traditional
	\textit{Conv-BN-LReLU} triplet.
		\begin{figure}[H]
		\includegraphics[width=0.49\linewidth]{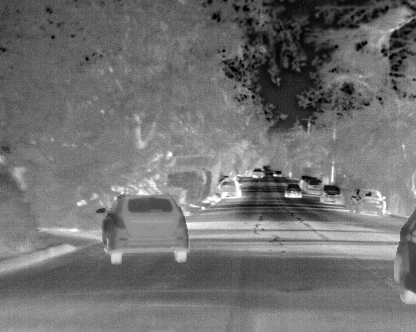}	
		\includegraphics[width=0.49\linewidth]{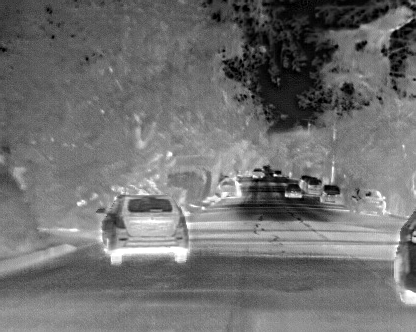}\\
		\includegraphics[width=0.49\linewidth]{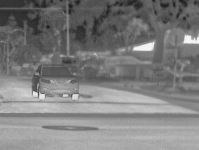}	
		\includegraphics[width=0.49\linewidth]{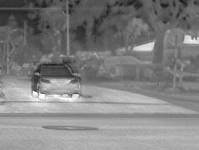}
		\caption{GAN translation synth to real sample (zoomed in and cropped). \textit{Left:} Original synthethic image from the simulator; \textit{Right:} GAN translated version. The GAN learns to transfer the heat emitted from the bottom of a car, a feature missing from the simulator.} \label{scenes_ganned}
	\end{figure}
	Similarly to \cite{mo2019instagan} the model is fed both the input image and the segmentation mask (see Fig.~\ref{scenes_2}) as to provide guidance to the model for the different thermal signatures of each category. Fig.~\ref{scenes_ganned} provides a sample of the synthetic to real thermal translation; it is interesting to note how the GAN has created the thermal reflection of the heat of the engine on the concrete of the road, that was not part of the 3D model. This physical property has been learned by the GAN from the FLIR thermal dataset.  Differently from \cite{mo2019instagan} we do not want our segmentation mask to be modified by the generative model as this would change its bounding box and be detrimental to the detector performances.

	\subsection{Other data augmentations}
	We have also evaluated the applicability of data augmentation techniques developed for RGB images in the thermal spectrum.

	\subsubsection{RandAugment}
\begin{figure}[H]
		\centering
		\includegraphics[width=0.49\linewidth]{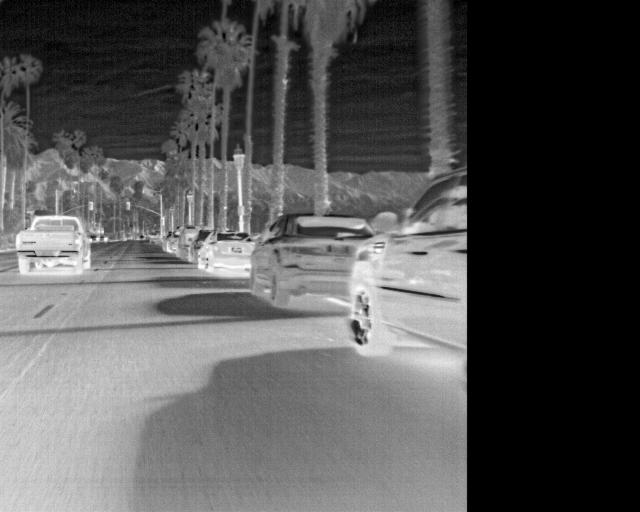}
		\includegraphics[width=0.49\linewidth]{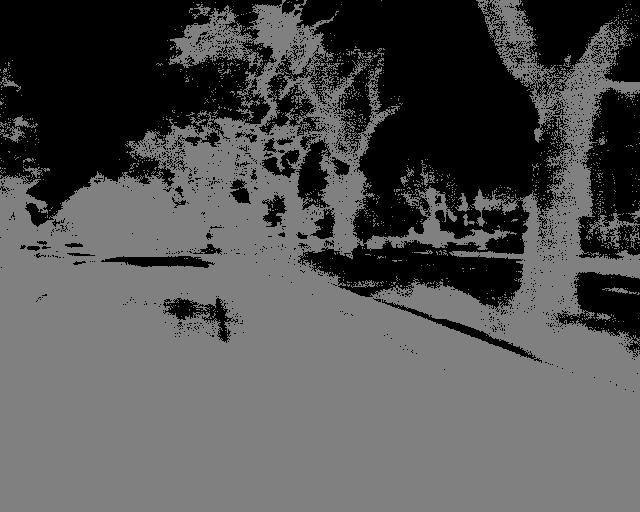}
		\caption{Examples of augmentations obtained with RandAugment \cite{randaug}. \textit{Left)} translateX and autocontrast have been applied to a picture of the FLIR dataset; \textit{Right)} in this case the pedestrians, bicycles and cars are not detectable at the all due to the solarization transformation.}\label{fig:randaugment}
	\end{figure}

	We searched the best combinations and strengths of augmentations over a subset of the FLIR training dataset composed of 5,000 images.
	However, we noticed that some of the transformations used are not suitable for the thermal domain, as shown in Fig.~\ref{fig:randaugment}.
	
	\subsubsection{BBox Augmentation}
\begin{figure}[H]
		\centering
		\includegraphics[width=0.49\linewidth]{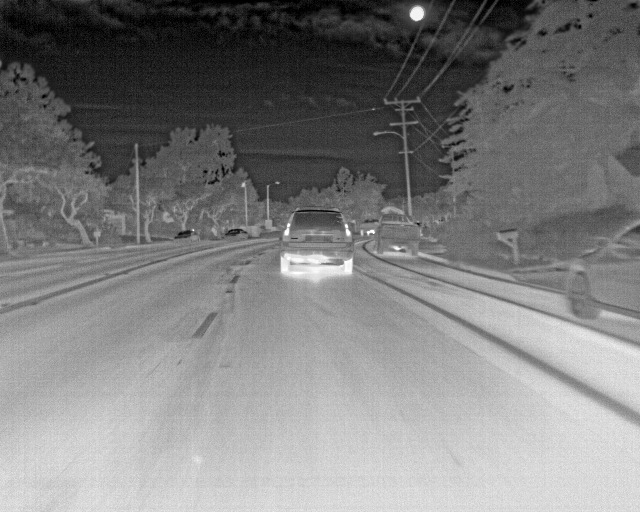}	
		\includegraphics[width=0.49\linewidth]{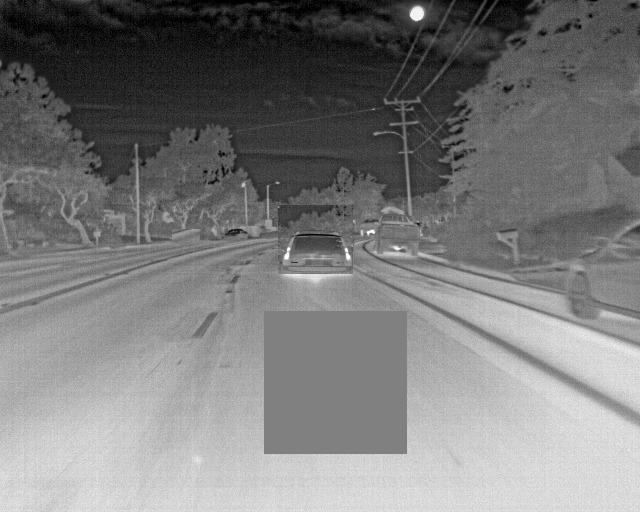}
		\caption{Left) An image from FLIR dataset; Right) BBox Augmented version. The content of the box's car in the middle is translated downwards and a cutout square has been added.
		} 
		\label{fig:bbox}
	\end{figure}

	While data augmentation methods like RandAug are often used for image classification models, Boundary-Box Augmentation \cite{zoph2020learning} (BBAug) focuses on object detection. In the FLIR dataset the annotations refer to objects like bikes, pedestrians and cars; the bounding box annotations of these objects open up the possibility to modify the content of the boxes. The transformations include, for example, to keep the same image while translating or zooming the content of the boxes. Starting from an original picture of the FLIR data set in figure \ref{fig:bbox}, the content of the bounding box in the middle is modified by this technique.
	
	BBAug \cite{zoph2020learning} is based on the concept of policy, an unordered set of K sub-policies. Each sub-policy is a set of transformations to be applied to the image, similarly to the AutoAugment method originally proposed in \cite{autoaug}.
	The authors provide the best sub-policies using a Reinforcement Learning search space algorithm. The models are trained on COCO data set with ResNet-50 backbone \cite{res-net} and RetinaNet detector \cite{retina-net}. For each picture we want to augment, a sub-policy is randomly selected. This method allows us to skip the very expensive search space phase.
	
	\subsection{Object detector}
	We used the layer-wise thermal adaptation of YOLOv3 originally proposed in \cite{kieu2019domain} as object detector, as a baseline for the experiments and to evaluate the data augmentation technique proposed in this paper.

	\section{Experimental results}
	\subsection{Dataset and experimental setup}
	All  of  our  experiments  were  conducted  on  the FLIR ADAS dataset \cite{flir-adas}.
	The dataset consists of over 10K images, collected using the FLIR Tau2 camera, both during the day (60\%) and the night (40\%) driving on Santa Barbara, CA with clear to overcast weather.
	We used the three most common classes offered by the FLIR-ADAS dataset, composed by
	Person (28,151), Car (46,692), and Bicycle (4,457), as this is the commonly used experimental setup and allows us to compare the proposed method with other state-of-the-art approaches.
	
	The training set contains 8,862 frames, while the test set contains 1,366 images taken on different streets from the training set ones. Several detectors have been compared using the mAP score.
	For the GAN training we used a subset of the FLIR-ADAS data from \cite{zhang2020multispectral} which provides pixel-aligned couples of RGB and thermal frames. We then used the RGB frames to extract segmentation masks for pedestrians and vehicles from the scene, using a pretrained on MSCOCO MaskRCNN \cite{he2017mask} from Detectron2 \cite{Detectron2018}.
	
	We devise and test multiple augmentation strategies by combining the different source of data augmentation previously discussed. From here on we will refer to BBox Augmentation \cite{zoph2020learning} as ``BBAug" and to we will keep the name RandAugment for the method of \cite{randaug}.
	
	As previously noted, we added synthetic data to the training set. This data is split into multiple sets in order to perform ablation studies:
	\begin{itemize}
		\item  Synth${_a}$ : Composed of pedestrians walking on a railroad scene; No cars and no roads. About 4K samples.
		\item  Synth${_b}$ : Composed of both cars and pedestrians over FLIR-ADAS training set scenes. About 10K samples.
		\item  Synth${_c}$ :  Composed of both cars and pedestrian over a railroad scene; no roads.
		About 4K samples.
	\end{itemize}
	
	We will use the following notation: GAN${_x}\rightarrow{_y}$ to refer to an experiment in which a generative model is trained over the subset  Synth${_x}$, performs inference over Synth${_y}$ and finally this  \textit{translated} Synth${_y}$ data is added to the detector training set.
	We also tested the generative model from \cite{kieu2021robust} which performs RGB $\rightarrow$ Thermal translation, and refer to it as GAN${_{rgb}}\rightarrow{_{th}}$.


	\subsection{Ablation studies}
	In the first experiment, we investigated several data augmentation strategies to improve the performances of our detector. In Table \ref{tab:augmentations} we compare the results of different augmentation techniques and the baseline from \cite{kieu2019domain} in terms of mean average precision (mAP) for all the detected classes. In the second and third experiment, we tested RandAugment \cite{randaug} and BBAug \cite{zoph2020learning}, by picking the best policy for each (see Tab.~\ref{tab:bbaug_policies}) and the best amount of augmented data (see. Tab.~\ref{tab:percent}). 
	
	Analyzing the results of Tab.~\ref{tab:augmentations} we can observe that BBAug \cite{zoph2020learning} performed better than Randaugment, bringing the base detector mAP to 75\%.
	Employing Randaugment \cite{randaug} instead does not show much improvement, probably due to the fact that its transformations are not developed for object detection as shown in Fig.\ref{fig:randaugment}.
	The combination of the aforementioned does note give better results than BBAug \cite{zoph2020learning} alone.
	
	The augmentation strategy using the generative model from \cite{kieu2021robust} provides some improvements compared to the baseline method but only by a 0.3\% mAP. 
	
	The injection of synthetic data, through compositing of fake 3D objects in a thermal scene, instead, both directly from the simulator (Synth${_a}$, Synth${_b}$, Synth${_c}$) and after the domain translation (GAN${_b}\rightarrow{_a}$, GAN${_b}\rightarrow{_c}$), raises the detection performances across all the classes.
	The best pipeline we found working uses the translated synthetic data from our Synth${_a}$ subset, reaching a 76.4\% mAP. Also using the same subset followed by BBAug~\cite{zoph2020learning} improved the detector mAP performances to 75.6\% which is the second best result. The combination of these two last techniques does not yield better results.
	In general, the proposed idea of compositing fake 3D thermal objects in a thermal scene is an effective augmentation technique. This is further improved adding BBAug or GAN-based augmentation that improves the appearance of the 3D objects. Instead, combining both GAN and BBAug does not improve the results.


	\begin{table}[H]
		\caption{Comparison of different augmentation strategies in terms of mAP. Best in bold, second best underlined.}
		\centering
		\resizebox{\columnwidth}{!}{
			\begin{tabular}{ l r r r r}
				\hline
				\thead{Augmentation Strategy}    & \thead{Person} & \thead{Bicycle} & \thead{Car} & \thead{mAP} \\ \hline
				None (Baseline)~\cite{kieu2019domain}			& 75.6	& 57.4	& 86.5	& 73.2 \\ \hline
				RandAugment \cite{randaug}                    & 74.4                        & 60.2                         & 85.4                     & 73.3                     \\ 
				BBAug \cite{zoph2020learning}                         & 79.4                        & 58.4                         & 87.2                     & 75.0                     \\
				RandAugment~\cite{randaug} + BBAug~\cite{zoph2020learning}              & 74.6                        & 61.6                         & 86.0                     & 73.9                     \\ 
				GAN${_{rgb}}\rightarrow{_{th}}$ \cite{kieu2021robust} & 77.1                        & 56.9                         & 86.4                     & 73.5                     \\ 
				Synth${_a}$                   & 79.3                        & 60.1                         & 86.2                     & 75.2                     \\ 
				Synth${_a}$ + BBAug~\cite{zoph2020learning}                   & 77.1                        & 64.2                         & 85.6                     & \underline{75.6}            \\ 
				
				Synth${_b}$                   &       77.2                    &              60.0                  &          86.2                &        74.5                  \\ 
				Synth${_c}$                   &           75.9              &        60.5                        &               85.2          &          73.9              \\
				GAN${_b}\rightarrow{_c}$              &       77.6                  &  59.1                             &     85.6                    &        74.1   \\
				GAN${_b}\rightarrow{_a}$              &       78.5                    &  64.0                              &     86.9                      &        \textbf{76.4}             \\
				
				GAN${_b}\rightarrow{_a}$   + BBAug~\cite{zoph2020learning}      &       74.7                 &  64.1                             &     86.5                    &        75.1              \\ 
				\hline
				
			\end{tabular}
		}
		\label{tab:augmentations}
	\end{table}

	As previously explained, Boundary Box augmentation \cite{zoph2020learning} introduces 4 policies each composed of 2 possible transformations that can be applied to an image. We tested all of them and present them in Tab.~\ref{tab:bbaug_policies}, concluding that the policy v0 (i.e.~TranslateX $p=0.6$, $m=4$, Equalize $p=0.8$, $m=10$) is preferable in FLIR-ADAS. All the experiments in Tab.~\ref{tab:augmentations}, Tab.~\ref{tab:percent}, Tab.~\ref{tab:compare} use this policy. 
	\begin{table}[H]
		\caption{mAP for the 4 policies of the BBAug~\cite{zoph2020learning} over a subset of the synthetic data.}
		\centering
		\resizebox{\columnwidth}{!}{
			\setlength{\tabcolsep}{12pt}
			\begin{tabular}{l  r r r r }
				\hline
				Policies & 	\thead{v0}           & 	\thead{v1}   & 	\thead{v2}   & 	\thead{v3}   \\ \hline
				Person   & 77.1          & 73.4 & 72.2 & 72.9 \\ 
				Bicycle  & 64.2          & 53.7 & 54.3 & 54.5 \\ 
				Car      & 85.6          & 84.9 & 82.6 & 84.6 \\ 
				mAP      & \textbf{75.6} & 70.7 & 69.7 & 70.7 \\ \hline
			\end{tabular}
		}
		\label{tab:bbaug_policies}
	\end{table}
	
	We have also explored the effect of different quantities of augmented data w.r.t the original train set size. In Table \ref{tab:percent} a non exhaustive analysis of the percentage of fake w.r.t real suggests, as previously noted in \cite{kieu2021robust}, that the optimal amount of generated data to add is between 10\% and 20\%, for the tested strategies increasing the amount of augmented data becomes detrimental to learning.
	
	
	\begin{table}[H]
		\caption{Ablation study on varying quantities of augmented images. The percentage indicates the relative size of the added augmented data.	}
		\centering
		\resizebox{\columnwidth}{!}{
			\begin{tabular}{l r r r}
				\hline
				\thead{Technique}    & \thead{@10\%} & \thead{@20\%} & \thead{@50\%} \\ \hline
				BBAug~\cite{zoph2020learning}           & 74.8      & 75.0       & 74.4             \\ 
				RandAugment~\cite{randaug}          &   72.9 & 73.3     &   66.5      \\ 
				RandAugment~\cite{randaug} + BBAug~\cite{zoph2020learning}      & 73.9   & 73.7     & 73.6   \\ 
				GAN${_{rgb}}\rightarrow{_{th}}$~\cite{kieu2021robust}     &   73.5       &   72.6      &    72.7    \\ 
				Synth${_a}$   & 74.2 & 74.1 & 74.0  \\ \hline
			\end{tabular}
			
			\label{tab:percent}
		}
	\end{table}

	\subsection{Comparison with state-of-the-art approaches}
	In this experiment we compare the best augmentation techniques evaluated in the previous section, comprising our proposed composition of fake thermal 3D objects with and without GAN processing, with state-of-the-art approaches.
	All the following experiments are evaluated in terms of Mean Average Precision (mAP).
	
	\begin{table}[H]
		\caption{Comparison of the proposed method with state-of-the-art multispectral and thermal-only methods. Best in bold, second best underlined.}\label{tab:compare}
		\centering
		\resizebox{\columnwidth}{!}{
			\begin{tabular}{l l r r r r }
				\hline
				\thead{Technique} & \thead{Spectra} & \thead{Person} & \thead{Bicycle} & \thead{Car} & \thead{mAP} \\ \hline
				MMTOD-UNIT \cite{devaguptapu2019borrow}						& Multi	& 64.5	& 49.4	& 70.8	& 61.5 \\
				ABiFN \cite{charan2020abifn}				& Multi	& 66.1	& 48.5	& 71.8	& 62.1 \\
				SSD512 VGG16 \cite{munir2020thermal}		& Multi & 71.0	& 55.5	& 82.3	& 69.6 \\ 
				CFRM\_3 \cite{zhang2020multispectral}	& Multi & 74.5	& 57.8	& 84.9	& 72.4 \\ 
				SSTN101	\cite{munir2021sstn}				& Multi & -		& - 		& - 	& 77.6 \\ 
				\hline 
				YoloV3 Transfer\cite{gaus2020visible}	& Thermal	& 33.2	& 34.5	& 55.4	& 41.0 \\
				SSD VGG-16 Transfer \cite{gaus2020visible}	& Thermal	& 61.9	& 46.1	& 85.1	& 64.4 \\
				Layer-Wise \cite{kieu2019domain}			& Thermal	& 75.6	& 57.4	& 86.5	& 73.2 \\ 
				RefineDetect512 \cite{flir-adas}			& Thermal	& 79.4	& 58.0	& 85.6	& 74.3 \\ 
				ThermalDet \cite{9064036}				& Thermal	& 78.2	& 60.0	& 85.2	& 74.6 \\
				\hline
				BBAug \cite{zoph2020learning}			& Thermal	& 79.4  & 58.4  & 87.2  & 75.0 \\ 
				Synth${_a}$								& Thermal	& 79.3  & 60.1  & 86.2  & 75.2 \\ 
				Synth${_a}$  + BBAug\cite{zoph2020learning}	& Thermal	& 77.1	& 64.2  & 85.6  & \underline{75.6} \\ 
				GAN${_b}\rightarrow{_a}$		& Thermal	& 78.5	& 64.0	& 86.9	& \textbf{76.4} \\ \hline
			\end{tabular}
		}
	\end{table}
	
	Table~\ref{tab:compare} compares our results with the state-of-the-art single and multi-spectral approaches.
	We distinguish multi-spectral detectors, i.e.~models which at test time detect objects using both visible and thermal spectrum images, and thermal only detectors, i.e.~models that use only the thermal spectrum images. Models which used visible images for transfer learning only during training fall under the \textit{Thermal} category. 

	YoloV3 Transfer\cite{gaus2020visible} and SSD VGG-16 Transfer \cite{gaus2020visible} are both pretrained models on Coco and ImageNet data sets  and tested on FLIR-ADAS.
	As shown in the Tab.\ref{tab:compare}, our thermal only detector using different augmentations outperform all but one of the multi-spectral detectors. Only Self-Supervised Thermal Network (SSTN101) \cite{munir2021sstn} got better results with 77.57\% mAP. Instead, considering thermal-only detectors, our proposed method outperforms all the competing approaches.

	

\begin{figure*}[!htb]
		
		\textbf{Layer-wise - \cite{kieu2020layerwise}} \hspace{0.1\textwidth} \textbf{Synth${_a}$ - our method} \hspace{0.1\textwidth} \textbf{BBaug - \cite{zoph2020learning}}
		\centering
		\includegraphics[width=0.32\textwidth]{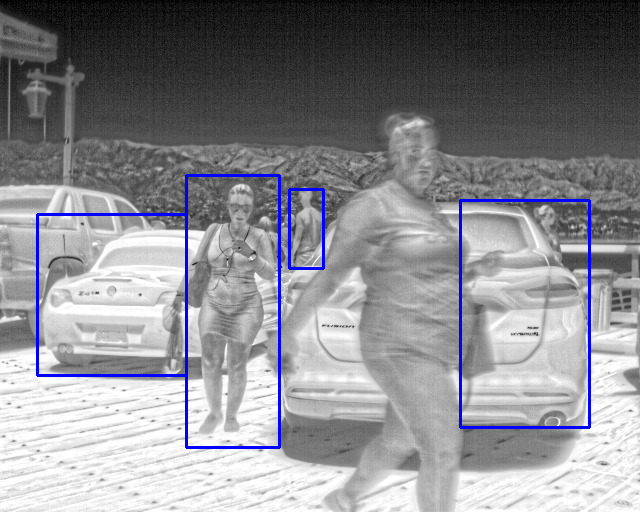}
		\includegraphics[width=0.32\textwidth]{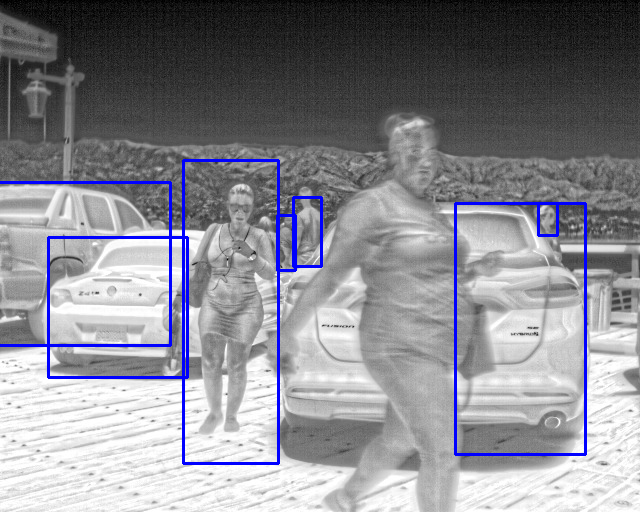}	
		\includegraphics[width=0.32\textwidth]{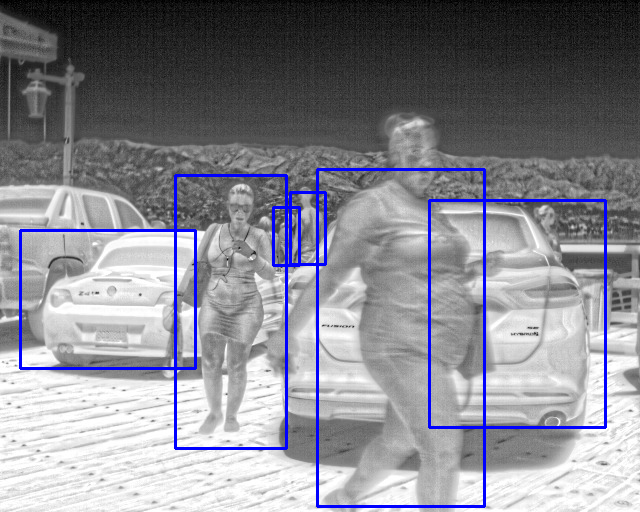}
		\includegraphics[width=0.32\textwidth]{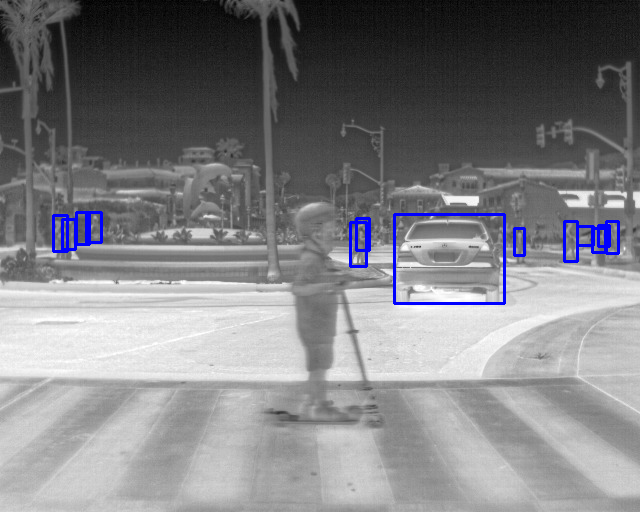}
		\includegraphics[width=0.32\textwidth]{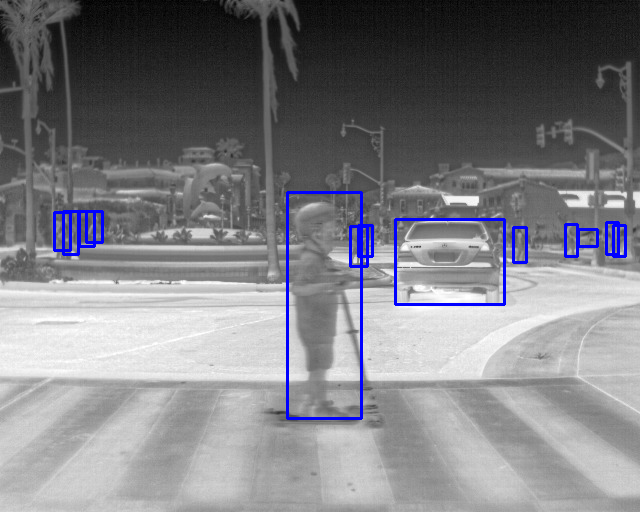}	
		\includegraphics[width=0.32\textwidth]{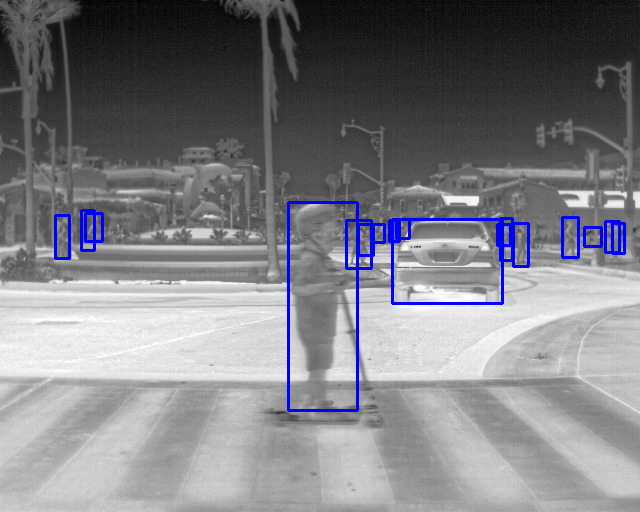}
		\includegraphics[width=0.32\textwidth]{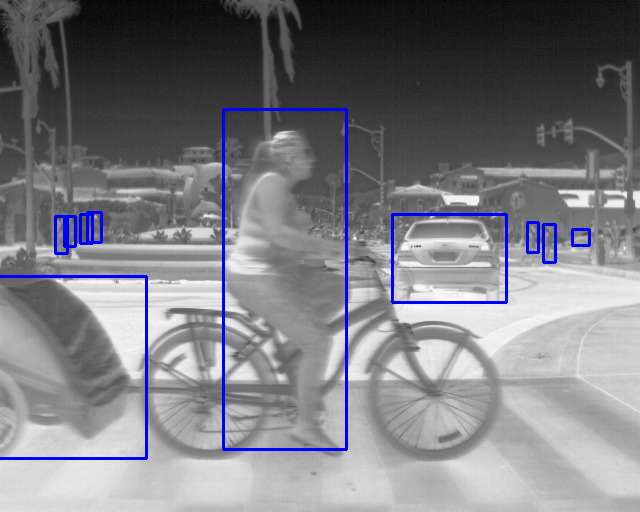}
		\includegraphics[width=0.32\textwidth]{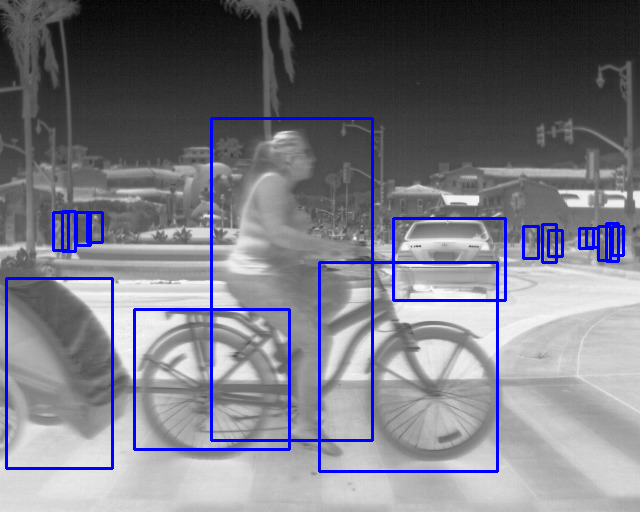}	
		\includegraphics[width=0.32\textwidth]{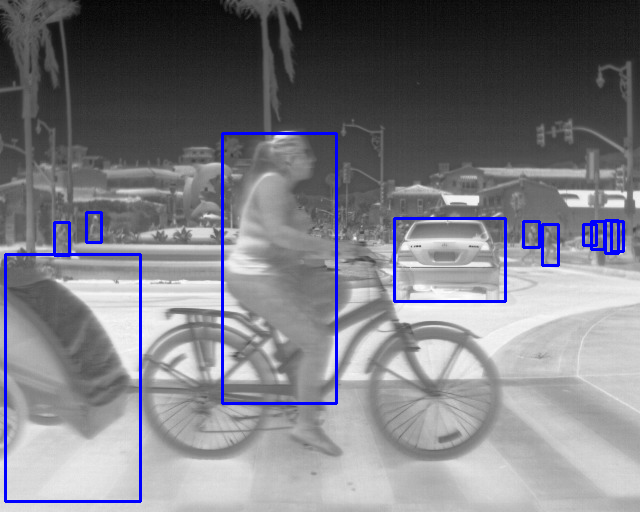}
		\includegraphics[width=0.32\textwidth]{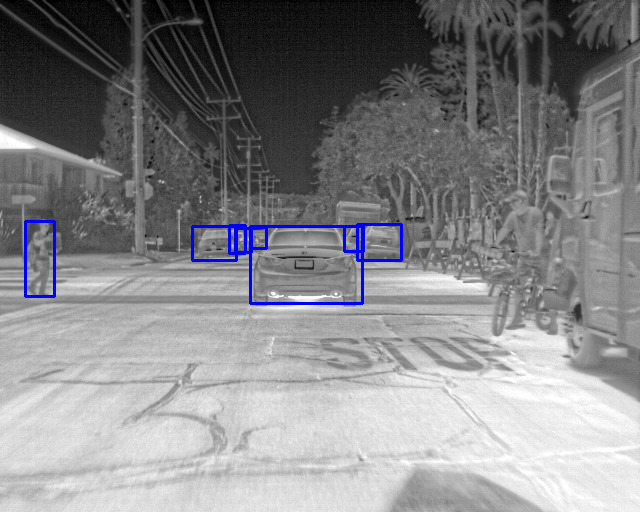}
		\includegraphics[width=0.32\textwidth]{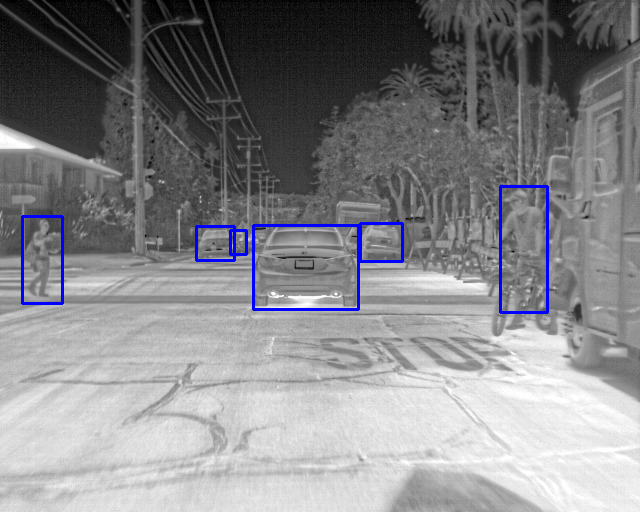}	
		\includegraphics[width=0.32\textwidth]{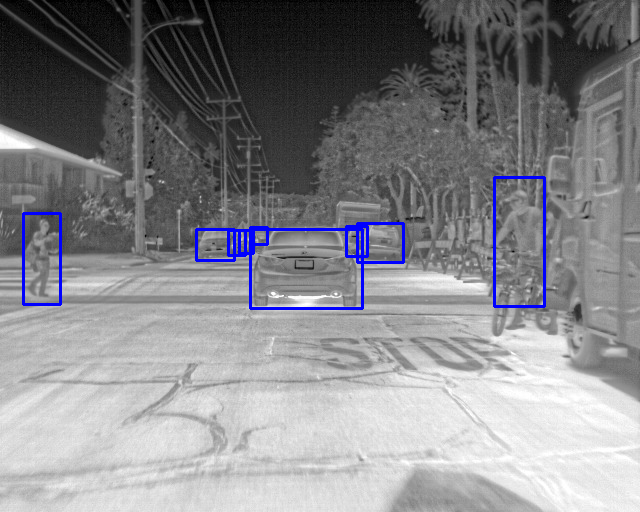}
		\caption{Examples of detections: \textit{left)} Layer-wise adaptation \cite{kieu2020layerwise}, \textit{middle)} our method, \textit{right)} BBaug augmentation. In general, augmentations help to detect more persons, even in very critical cases like the kid that is crossing the street on row 2. Our propose method helps the network to better distinguish bicycle from persons, as shown on row 3.} \label{fig:comparison}
	\end{figure*}

	Fig.~\ref{fig:comparison} compares the detections obtained using the layer-wise YOLOv3 baseline presented in \cite{kieu2020layerwise}, the results obtained adding the BBAug augmentation technique \cite{zoph2020learning} and the results obtained using our proposed approach, i.e.~adding 3D models and improving their appearance with a GAN. The examples on row 2 show that the kid crossing the street was not detected by the method proposed in \cite{kieu2020layerwise}, but both augmentation approaches are now capable of correctly detecting him. Row 3 shows how adding 3D models, as proposed in our method, help to distinguish persons from bicycles, that are not detected with the two other approaches. Another example of correct detection of a bicycle, is shown in row 4, were both augmentations are able to recognize it.
	
	\section{Conclusions}
	In this paper we compared several data augmentation strategies to improve a YOLOv3 detector working in a thermal-only domain. Given the challenges of the task, such as its data scarcity, we presented an augmentation pipeline that combines simulated data within a real scene, followed by domain adaptation using a generative model. This approach is particularly suitable for domains, such as thermal imagery, where the creation of completely synthetic scenes is unfeasible or extremely expensive due to the difficulty of modeling all the physical properties of the entities of the scene.
	Our best combination of strategies reaches states of the art performance in the thermal only domain, and reaches similar or better results with respect to multi-spectral detectors.

		We gratefully acknowledge the support of NVIDIA Corporation with the donation of the Titan X Pascal GPU used for this research.
		This project was partially funded by Leonardo SpA.

	\balance
	\bibliographystyle{ACM-Reference-Format}
	\bibliography{document}

\end{multicols}
\end{document}